%% file: plslam.tex
\documentclass[journal]{IEEEtran}
\IEEEoverridecommandlockouts
\usepackage{00_preamble}
\begin{document}
\input{01a_notation}

\input{01a_headers}
\input{1_introduction}

\input{2_related_work}

\input{3_system_overview}
\input{4_feature_tracking}
\input{5_mapping}
\input{6_loop_closure}
\input{7_experiments_2ndreview}

\input{8_conclusions}
\input{9_bib}
\end{document}

%% file: 01a_notation.tex

\newcommand{\ruben}[1]  {\textcolor{black}{#1}}	
\newcommand{\paco}[1]   {\textcolor{black}{#1}}		
\newcommand{\javier}[1] {\textcolor{black}{#1}}	

\newcommand{\review}[1] {\textcolor{black}{#1}}		

\newcommand{\toreview}[1] {\textcolor{red}{#1}}		

\newcommand{\tpaco}[1]  {\todo[inline, color=red!40  ]{\textbf{FM}: {\footnotesize \normalfont #1}}}
\newcommand{\tjavier}[1]{\todo[inline, color=green!40]{\textbf{JG}: {\footnotesize \normalfont #1}}}
\newcommand{\truben}[1] {\todo[inline, color=blue!40] {\textbf{RG}: {\footnotesize \normalfont #1}}}

\newcommand{\hastum}{01}    

\newcommand {\lbavar}   {\bs{\psi}}				
\newcommand {\kfgroup}  {\mathcal{K}}			
\newcommand {\ptgroup}  {\mathcal{P}}			
\newcommand {\lsgroup}  {\mathcal{L}}			
\newcommand {\kfgroupL} {\mathcal{K}_{l}}		
\newcommand {\ptgroupL} {\mathcal{P}_{l}}		
\newcommand {\lsgroupL} {\mathcal{L}_{l}}		
\newcommand {\kf}[1] 	{\separam_{#1 w}}		
\newcommand {\pt}[1]    {\textbf{X}_{w #1}}		
\newcommand {\lsP}[1]   {\textbf{P}_{w #1}}		
\newcommand {\lsQ}[1]   {\textbf{Q}_{w #1}}		
\newcommand {\ptL}[1]    {\textbf{X}_{#1}}		
\newcommand {\lsPL}[1]   {\textbf{P}_{#1}}		
\newcommand {\lsQL}[1]   {\textbf{Q}_{#1}}		
\newcommand {\ptproj}[1]    {\textbf{x}_{#1}}	
\newcommand {\lsproj}[1]    {\textbf{l}_{#1}}	
\newcommand {\lsPproj}[1]   {\textbf{p}_{#1}}	
\newcommand {\lsQproj}[1]   {\textbf{q}_{#1}}	
\newcommand {\projfun}[1]    {\bs\pi{#1}}		
\newcommand {\ptcolor}{black}
\newcommand {\lscolor}{black}

\newcommand {\indfig}{-4mm}						
\newcommand {\indtab}{-4mm}						
\newcommand {\indsubfig}{0mm}

\newcommand{\wover}      {\columnwidth}				
\newcommand{\woptimer}   {0.30\textwidth}				
\newcommand{\wresults}   {0.43\textwidth}				
\newcommand{\wlegend}    {0.30\textwidth}				
\newcommand{\wkitti}     {0.21\textwidth}				
\newcommand{\wtsukuba}   {0.235\textwidth}				
\newcommand{\wtsukubafig}{0.215\textwidth}				

\newcommand{\wteaser}    {0.21\textwidth}
\newcommand{\wconfmat}   {0.48\textwidth}			
\newcommand{\wprcurves}  {0.48\textwidth}	
\newcommand{\wkittimap}  {0.47\textwidth}	
\newcommand{\wkittitraj} {0.30\textwidth}
\newcommand{\wtummap}    {0.21\textwidth}
\newcommand{\weurocgraph}{0.40\textwidth}
\newcommand{\weurocmapa} {0.22\textwidth}	
\newcommand{\weurocmapb} {0.20\textwidth}	

\newcommand{\fig}[1]{Figure \ref{#1}}					
\newcommand{\figs}[2]{Figures \ref{#1} and \ref{#2}}	
\newcommand{\tab}[1]{Table \ref{#1}}					
\newcommand{\secref}[1]{Section {\ref{#1}}}				

\newcommand{\bs}[1]{\boldsymbol{#1}}					
\newcommand{\ssl}[1]{\tensor[^{#1}]}					
\newcommand{\MatrixS}[1]{\bs{#1}}						
\newcommand{\MatrixL}[1]{\textbf{#1}}					
\newcommand{\brackets}[1]{\begin{bmatrix}#1\end{bmatrix}}		

\newcommand{\argmin}[1]{\underset{#1}{\operatorname{argmin}}}		
\newcommand{\argmax}[1]{\underset{#1}{\operatorname{argmax}}}		
\newcommand{\der}[2]{\frac{\partial #1}{\partial #2}}			
\newcommand{\derin}[3]{\left.\der{#1}{#2}\right|_{#3}} 			
\newcommand{\prob}[2]{p\left( #1 | #2 \right)}				
\newcommand{\norm}[1]{\left\lVert#1\right\rVert} 			
\newcommand{\fnorm}[1]{\left\lVert#1\right\rVert_{\mathfrak{F}}}	
\newcommand{\skewmat}[1]{ \left[#1\right]_\times }			

\newcommand{\IdMat}{\MatrixL{I}}					
\newcommand{\canonicalvec}[1]{\textbf{e}_{#1}}		
\newcommand{\TRANSPOSE}{^\top}						
\newcommand{\symcov}{\MatrixS{\Sigma}}				
\newcommand{\symRe}{\mathbb{R}}						
\newcommand{\symSSpace}{S}						    
\newcommand{\symPSpace}{\mathbb{P}}					
\newcommand{\symRotSpace}{SO(3)}					
\newcommand{\symRotLie}{\mathfrak{so}(3)}			
\newcommand{\symEucSpace}{SE(3)}					
\newcommand{\symEucLie}{\mathfrak{se}(3)}			
\newcommand{\symrot}{\textbf{R}}					
\newcommand{\symtrans}{\textbf{t}}					

\newcommand{\idL}{L}				
\newcommand{\idR}{R}				
\newcommand{\idF}{k}				
\newcommand{\idFn}{k+1}				

\newcommand{\lIm}{\textbf{l}}				
\newcommand{\normL}{\eta_l}					
\newcommand{\lImFirst}{\lIm_{\idL,\idF}}	
\newcommand{\lImSecond}{\lIm_{\idR,\idF}}	
\newcommand{\lImThird}{\lIm_{\idL,\idFn}}	
\newcommand{\lImFourth}{\lIm_{\idR,\idFn}}	

\newcommand{\match}{\textbf{m}}				

\newcommand{\cam}{C}						
\newcommand{\calib}{\MatrixL{K}}			
\newcommand{\LO}{LO}						
\newcommand{\spoint}{\textbf{p}}			
\newcommand{\epoint}{\textbf{q}}			
\newcommand{\Spoint}{\textbf{P}}			
\newcommand{\Epoint}{\textbf{Q}}			
\newcommand{\spointx}{p_x}					
\newcommand{\spointy}{p_y}					

\newcommand{\separam}{\bs{\xi}}
\newcommand{\separaminc}{\bs{\varepsilon}}
\newcommand{\separamopt}{\separam^*}
\newcommand{\reltrans}{\MatrixL{T}(\separam)}
\newcommand{\reltransopt}{\MatrixL{T}(\separamopt)}

\newcommand{\symover}{\gamma}
\newcommand{\errfun}{\MatrixL{E}}
\newcommand{\weifun}{\MatrixL{W}}
\newcommand{\jacfun}{\MatrixL{J}}

\newcommand{\cross}{\times}	
\newcommand{\lx}{(p_y-q_y)}
\newcommand{\ly}{(q_x-p_x)}
\newcommand{\lbeta}{(p_x^2+p_y^2+q_x^2+q_y^2Id)-2(p_xq_x+p_yq_y}

%% file: 01a_headers.tex
\title{\LARGE \bf 
PL-SLAM: a Stereo SLAM System through the Combination of Points and Line Segments
}


\author{Ruben Gomez-Ojeda, David Zu\~niga-No\"el, Francisco-Angel Moreno, \\ Davide Scaramuzza,
	and Javier Gonzalez-Jimenez
	\thanks{This work has been supported by the Spanish Government (project \paco{DPI2017-84827-R} and grant BES-2015-071606) and the Andalusian Government (project TEP2012-530).}
	\thanks{ R. Gomez-Ojeda, F.A. Moreno, D. Zu\~niga-No\"el, and J. Gonzalez-Jimenez are with the Machine Perception and Intelligent Robotics (MAPIR) Group, University of Malaga. (email: rubengooj@gmail.com).} 
	\thanks{ D. Scaramuzza is with the Robotics and Perception Group, Dep. of Informatics, University of Zurich, and Dep. of Neuroinformatics, University of Zurich and ETH Zurich, Switzerland.}} 






\maketitle

\begin{abstract}
Traditional approaches to stereo visual SLAM rely on point features to estimate the camera trajectory and build a map of the environment. 
In low-textured environments, though, it is often difficult to find a sufficient number of reliable point features and, as a consequence, the performance of such algorithms degrades.
This paper proposes PL-SLAM, a stereo visual SLAM system that combines both points and line segments to work robustly in a wider variety of scenarios, 
particularly in those where point features are scarce or not well-distributed in the image.
PL-SLAM leverages both points and segments at all the instances of the process: visual odometry, keyframe selection, bundle adjustment, etc. We contribute also with a loop closure procedure through a novel bag-of-words approach that exploits the combined descriptive power of the two kinds of features.
Additionally, the resulting map is richer and more diverse in 3D elements, which can be exploited to infer valuable, high-level scene structures like planes, empty spaces, ground plane, etc. (not addressed in this work).
Our proposal has been tested with several popular datasets (such as KITTI and EuRoC), and is compared to state of the art methods like ORB-SLAM, revealing a more robust performance in most of the experiments, while still running in real-time.
An open source version of the PL-SLAM C++ code will be released for the benefit of the community.
\end{abstract}




%% file: 1_introduction.tex


\section{Introduction}
\label{sec_introduction}
\begin{figure}[!htb]
	\begin{center}
		\centering
		\subfigure[\textit{lt-easy}]{	
			\includegraphics[width=\wteaser]{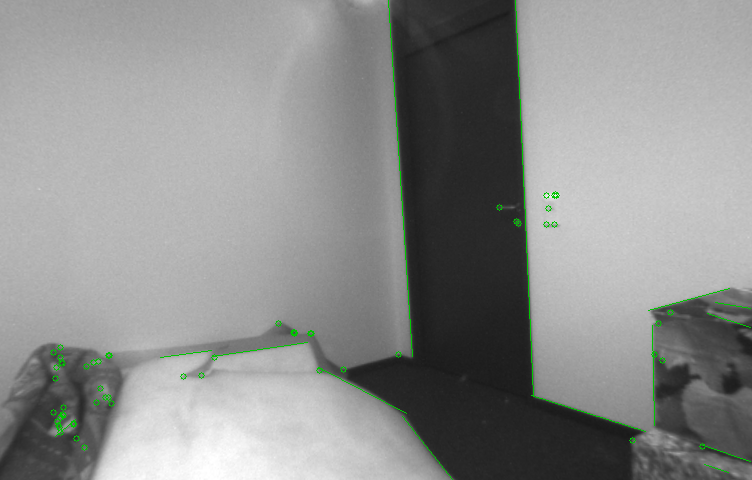}
			\vspace{\indfig}
			\label{lt-zurich}
		}	
		~
		\subfigure[\textit{euroc/V2-01-easy}]{	
			\includegraphics[width=\wteaser]{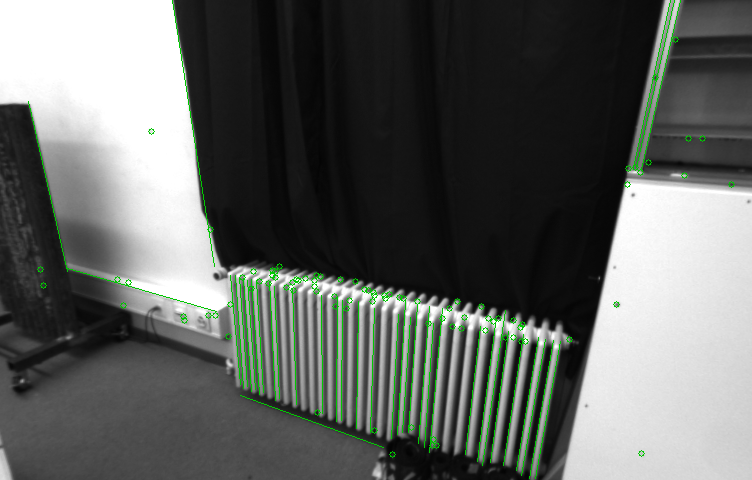}
			\vspace{\indfig}
			\label{lt-euroc}
		}		
		~	
		\subfigure[\textit{euroc/V1-01-easy}]{	
			\includegraphics[width=\wteaser]{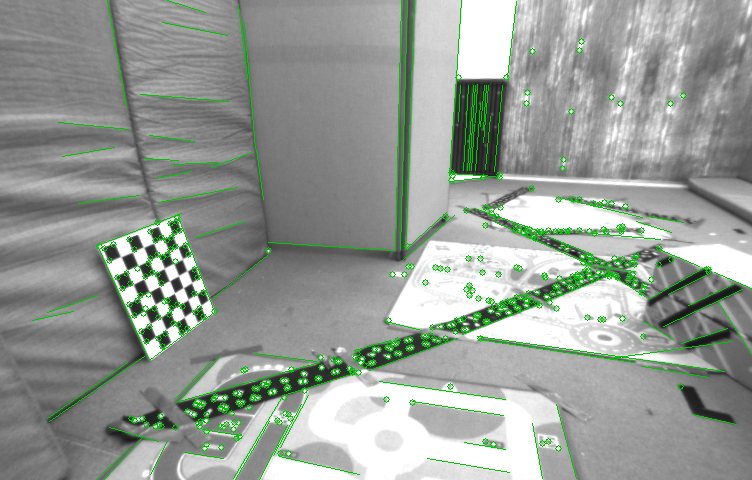}
			\vspace{\indfig}
			\label{euroc-obs}
		}	
		~
		\subfigure[Map from \subref{euroc-obs}]{	
			\includegraphics[width=\wteaser]{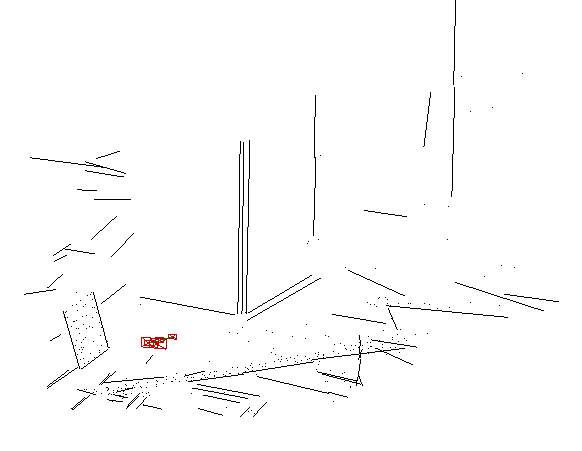}
			\vspace{\indfig}
			\label{euroc-map}
		}	
		~
		\caption{
		\ruben{
		Low-textured environments
		are challenging for \textit{feature-based} SLAM systems 
		based on traditional keypoints.	
		In contrast, line segments are usually common in human-made environments, and apart from an improved 
		camera localization, the built maps are richer as they are populated with more meaningful elements (3D line-segments).
		}
		}
		\vspace{\indfig}
		\vspace{\indfig}
		\label{fig_feature_tracking}
	\end{center}
\end{figure}

In recent years, visual Simultaneous Localization And Mapping (SLAM) is firmly progressing towards the degree of reliability required for fully autonomous vehicles: mobile robots, self-driving cars or Unmanned Aerial Vehicles (UAVs).
%
%
In a nutshell, the SLAM problem consists of the estimation of the vehicle trajectory given as a set of poses (position and orientation), while simultaneously building a map of the environment. Apart from self-localization, a map becomes useful for obstacle avoidance, object recognition, task planning, etc. \cite{Durrant-Whyte2006a}. 

As a first-level classification, SLAM systems can be divided into \textit{topological} (e.g. \cite{Milford2004,Cummins2008,Milford2012,Milford2013a}) and \textit{metric} approaches. In this paper, we focus on the latter, which take into account the \textit{geometric} information of the environment and build a physically meaningful map of it \cite{klein09cameraphone,Blanco2015}. These approaches can be further classified into \textit{direct} and \textit{feature-based} systems.
%
\ruben{	
The first group, i.e. \textit{direct} methods, estimates the camera motion  by minimizing the photometric errors between consecutive frames 
under the assumption of constant brightness along the local parts of the sequences 
(examples of this approach can be found elsewhere \cite{newcombe2011dtam,engel2014lsd,forster2016svo}).
While this group of techniques has the advantage of working directly with the input images regardless of any 
intermediate representation, they are very sensitive to brightness changes 
(this phenomena was addressed in \cite{engel2018direct}) and constrained to narrow baseline motions.
}
\ruben{
In contrast, \textit{feature-based} methods employ an indirect representation of the images, typically in the form 
of point features, that are tracked along the successive frames and then employed for recovering the pose by 
minimizing the projection errors \cite{scaramuzza2011visual,murTRO2015}.
%
%
}

It is noticeable that the performance of any of the above-mentioned approaches usually decreases in low-textured environments 
in which it is typically difficult to find a large set of keypoint features. The effect in such cases is an accuracy impoverishment and, occasionally, the complete failure of the system.
Many of such low-textured environments, however, contain planar elements that are rich in linear shapes, so it would be possible to extract line segments from them.
We claim that these two types of features (keypoints and segments) complement each other and its combination leads to a more versatile, robust and stable SLAM system.
Furthermore, the resulting maps comprising both 3D points and segments provide more structural information from the environment than point-only maps, as can be seen in the example shown in Figure \ref{fig_feature_tracking}\subref{euroc-map}. 
Thus, applications that perform high-level tasks such as place recognition, semantic mapping 
or task planning, among others, can significantly benefit from the richer information that can be inferred from them. 

These benefits, though, come at the expense of a higher computational burden in both detecting and matching line-segments in images \cite{bartoli2005structure}, and also in dealing effectively with segment-specific problems like partial occlusions, line disconnection, etc. {which complicate feature tracking and matching as well as the residual computation for the map and pose optimization.}
%
{Such} hurdles are the reason why the number of solutions that have been proposed in the literature to SLAM or Structure from Motion (SfM) with line features (e.g.  \cite{mei2006fast,Sola2009,zhang2011hand,hofer2015line3d,briales2016minimal}) is so limited.
Besides, the few solutions we have found only perform robustly in highly structured environments while showing unreliable results when applied to more realistic ones such as those recorded in the KITTI or EuRoC datasets.
%
{In this work, we address the segment-specific tracking and matching issues by discarding outliers through the comparison of the length and the orientation of the line features, while, for the residual computation, we represent segments in the map with their endpoints coordinates.}
%
Thus, the residuals between the observed segments and their corresponding lines in the map are computed by the distance between the projections of those endpoints on the image plane and the infinite lines associated to the observed ones. 
{This way, we are able to build a consistent cost function that seamlessly encompasses both point and line features.}
%
%

These two kinds of features are also employed to robustly detect loop closures during robot navigation, following a new bag-of-words approach that combines the advantages of using each of them to perform place recognition.
%
%
In summary, we propose a novel and versatile stereo visual SLAM system, coined PL-SLAM, which builds upon our previous Visual Odometry approach presented in \cite{gomez2016robust}, and combines both point and line segment features to perform real-time robot localization and mapping. The main contributions of this work are:
\begin{enumerate}[$\circ$]

\item \review{The first open source stereo SLAM system that employs point and line segment features in real time, hence being capable of operating robustly in low-textured environments where traditional point-only approaches tend to fail, while obtaining similar accuracy in the rest of the scenarios. Because of the consideration of both kinds of features, our proposal also produces rich geometrical maps.}
\review{
\item A new implementation of the bundle adjustment process that seamlessly accounts for both kinds of features while refining the poses of the keyframes. 
}
\item An extension of the bag-of-words approach presented in \cite{galvez2012bags} that takes into account the description of both points and line segments to improve the loop-closure process.
\ruben{
}
\end{enumerate}
A set of illustrative videos showing the performance of proposed system and an open source version of the developed C++ PL-SLAM library are publicly available at  \href{http://mapir.uma.es/mapirwebsite/index.php/people/164-ruben-gomez}{http://mapir.uma.es}
and \href{https://github.com/rubengooj/pl-slam}{https://github.com/rubengooj/pl-slam}.

%% file: 2_related_work.tex
\section{Related Work}
\label{sec_related}

Feature-based SLAM is traditionally addressed by tracking keypoints along successive frames and then minimizing some error function (typically based on re-projection errors) to estimate the robot poses \cite{slam16cadena}.
Among the most successful proposals we can highlight FastSLAM \cite{montemerlo2002fastslam}, PTAM \cite{klein2007parallel}\cite{klein2008improving}, SVO \cite{forster2014svo}\cite{forster2016svo}, 
and, more recently, ORB-SLAM  \cite{murTRO2015}, %
which relies on a fast and continuous tracking of ORB features \cite{rublee2011orb}, and a local bundle adjustment step with the continuous observations of the point features. 
%
However, all of the previous approaches tend to fail or reduce their accuracy in low-textured scenarios where the lack of repeatable and reliable features usually hinders the feature tracking process.
%
%
%
%
In the following, we review the state of the art of SLAM systems based on alternative image features to keypoints: i.e. edgelets, lines, or line segments.

One of the remarkable approaches that employs \textit{line} features is the one in \cite{Smith2006}, where the authors propose an algorithm to integrate them into a monocular Extended Kalman Filter SLAM system (EKF-SLAM).
In the refered paper, the line detection relies on an hypothesize-and-test method that connects several near keypoints to achieve real-time performance.
%
\textcolor{black}{
Other works employ \textit{edge} landmarks as features in monocular SLAM, as the one reported in \cite{Eade2009}, which does not only include the information of the local planar patch as in the case of keypoints, but also considers local edge segments, hence introducing new valuable information as the orientation of the so-called \textit{edgelets}.
In that work they derive suitable models for those kinds of features and use them within a particle-filter SLAM system, achieving nearly real-time performance.
More recently, authors in \cite{forster2016svo} also introduced edgelets in combination with intensity corners in order to improve robustness in environments with little or high-frequency texture.
}
%

A different approach, known as \textit{model-based}, incorporates prior information about the orientation of the landmarks derived from line segments.
Particularly, the method in \cite{zhang2011building} presents a monocular 2D SLAM system that employs vertical and horizontal lines on the floor as features for both motion and map estimation.
For that, they propose two different parameterizations for the vertical and the horizontal lines: vertical lines are represented as 2D points on the floor plane (placed the intersection point between the line and such plane), while horizontal lines are represented by their two end-points placed on the floor. Finally, the proposed models is incorporated into an EKF-SLAM system.
Another model-based approach is reported in \cite{Zhou2015a}, where 
%
the authors introduce structural lines in an extension of a standard EKF-SLAM system. The dominant directions of the lines are estimated by computing their vanishing points under the assumption of a Manhattan world \cite{coughlan1999manhattan}.
%
%
All these model-based approaches, though, are limited to very structured scenarios and/or planar motions, as they rely solely on line features.

The works in \cite{Sola2009,Sola2011} address a generic approach that compares the impact of eight different landmark parametrization for monocular EKF-SLAM, including the use of point and line features.
Nevertheless, such systems are only validated through analytic and statistical tools  
%
that assumed already known data association and that, unlike our proposal, do not implement a complete front-end that detect and track the line segments.
%
Another technique for building a 3D line-based SLAM system has been proposed in the recent work \cite{zhang2015building}. 
For that, the authors employ two different representations for the line segments: the Pl{\"u}cker line coordinates for the initialization and 3D projections, and an orthonormal representation for the back-end optimization.
Unfortunately, neither the source code is available nor the employed dataset contain any ground-truth, therefore it has not been  possible to carry out a comparison against our proposal.
%

\review{
Recently, line segment features have also been employed for monocular pose estimation in combination with points, due to the bad-conditioned nature of this problem.
For that, in \cite{gomez2016pl} the authors extended the semi-direct approach in \cite{forster2014svo} with line segments. Thanks to this pipeline, line segments can be propagated efficiently throughout the image sequence, while refining the position of the end-points under the assumptions of high frame rate and very narrow-baseline.
}

\review{
Finally, by the time of the first submission of this paper, a work with the same name (PL-SLAM, \cite{pumarola2017pl}) was published extending the monocular algorithm ORB-SLAM to the case of including line segment features computed through the LSD detector \cite{von2010lsd}. Apart from being a monocular system (unlike our stereo approach), their proposal deals with line tracking and matching in an essentially different way: they propagate the line segments by their endpoints and then perform descriptor-based tracking, which increases the computational burden of ORB-SLAM. Besides this computational drawback, when working with features detected with the LSD detector, the variance of the endpoints becomes quite pronounced, specially in challenging illumination conditions or very low-textured scenes, making more difficult wide-baseline tracking and matching between line features in non-consecutive frames. Our PL-SLAM approach, in contrast, does not make any assumption regarding the position of the lines endpoints so that our tracking front-end allows to handle partially occluded line segments, endpoints variance, etc., for both the stereo and frame-to-frame tracking, hence becoming a more robust approach to point-and-line SLAM.
}

%

%% file: 3_system_overview.tex
\section{PL-SLAM Overview}
\label{sec_overview}
\begin{figure}[tb]
  \centering
    \includegraphics[width=\wover]{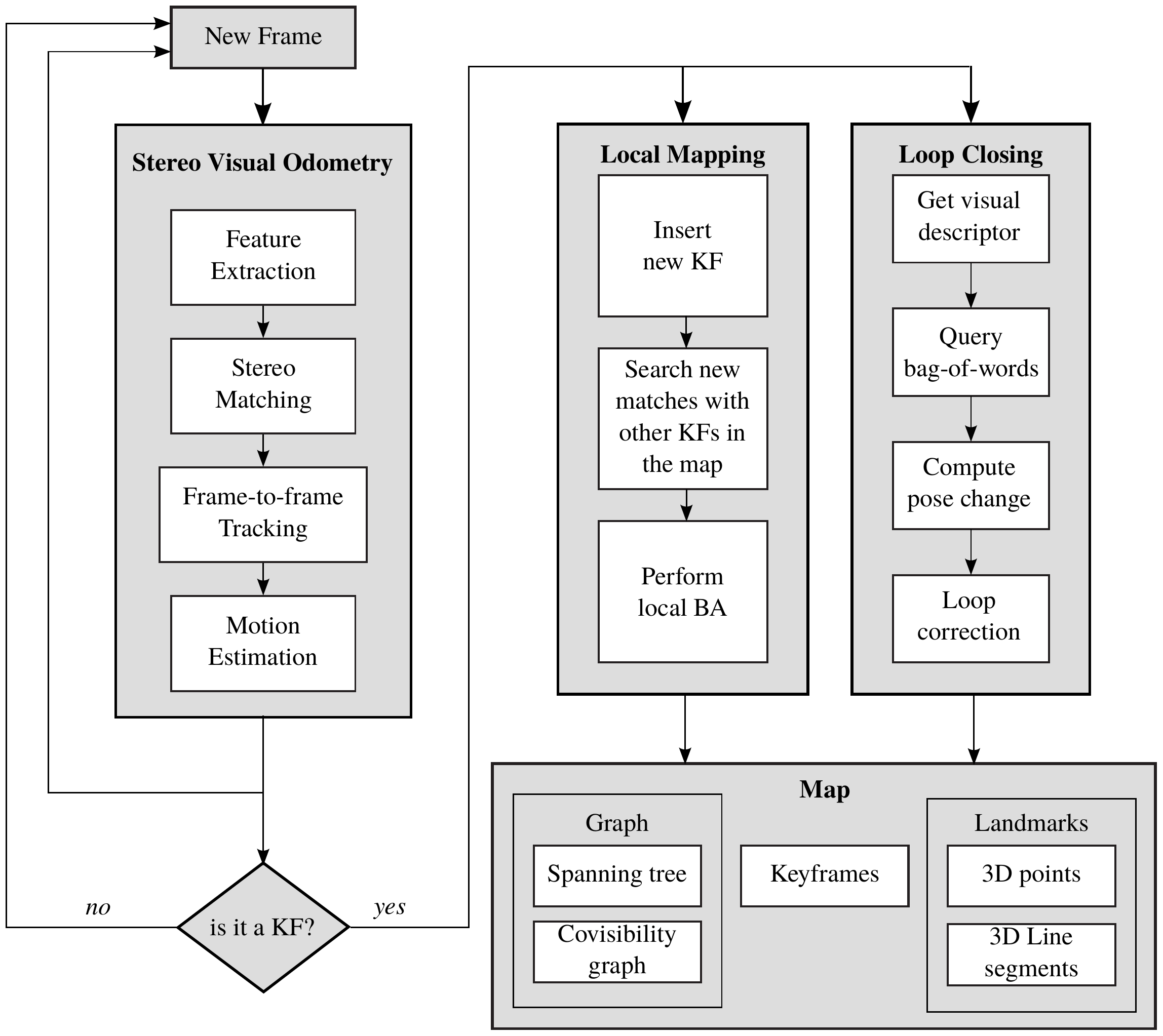}
    \caption{Scheme of the stereo PL-SLAM system. }
    \vspace{\indfig}
    \label{fig_system_overview}
\end{figure}

%
\review{
The general structure of the PL-SLAM system proposed here is depicted in \fig{fig_system_overview}, and its main modules are described in the following sections.
As it is common to other SLAM systems (being ORB-SLAM \cite{murTRO2015} the most popular method nowadays), our proposal is also based on three different threads: \textit{visual odometry}, \textit{local mapping}, and \textit{loop closure}.
This efficient distribution allows for a continuous tracking of the VO module while the local mapping and the loop closure ones 
are processed in the background only when a new keyframe is inserted.
}
%
%

\textbf{Map.} 
The map consists of i) a set of keyframes (KFs), ii) the detected 3D landmarks (both keypoints and line segments), iii) a covisibility graph and iv) a spanning tree.

The keyframes contain the observed stereo features and their descriptors, a visual descriptor of the corresponding left image computed through a visual vocabulary as explained later in \secref{sec_lc_detection}, and the information of the 3D camera pose.

Regarding the landmarks, we store the list of observations and the most representative descriptor for each detected landmark. Besides, specifically for points, we also keep its estimated 3D position while, for the line segments, we keep both their direction and the estimated 3D coordinates of their endpoints.

%
%
%

\begin{figure}[tb]
  \centering
    \includegraphics[width=\weurocgraph]{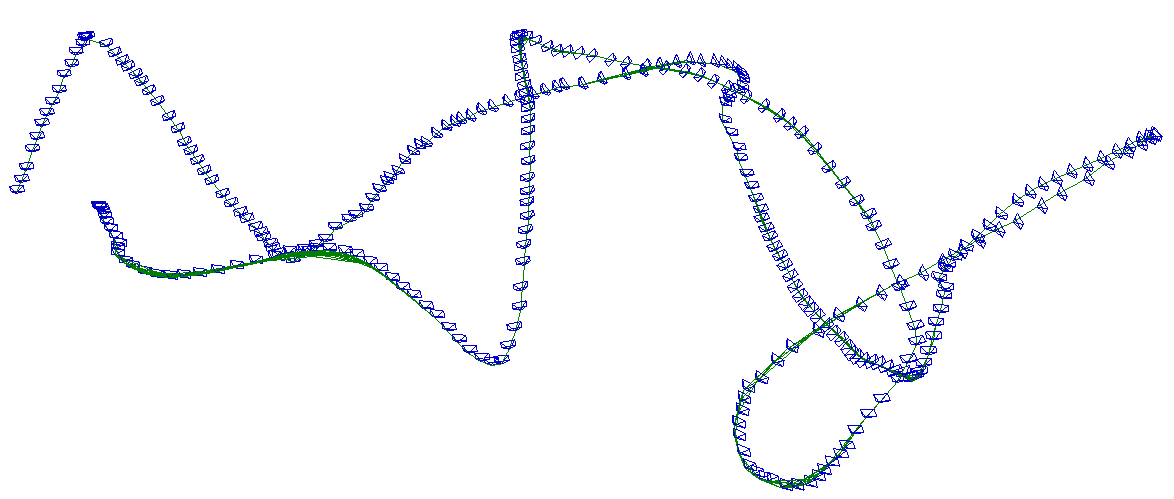}
    \caption{
    Covisibility graph in the sequence \textit{lt-first}
    for which we have represented the edges connecting the keyframes with green lines.  
    }
    \vspace{\indfig}
    \label{fig_euroc_graph}
\end{figure}

Finally, the covisibility information, as in \cite{strasdat2011double}, is modeled by a graph: each node represents a KF, and edges between KFs are created only if they share a minimum number of landmarks, which in this work is set to 20 landmarks (see \fig{fig_euroc_graph} for an example)\ruben{
, allowing for real-time bundle adjustment along the local map.
}

Similarly, in order to perform a faster loop closure optimization, we also form the so-called \textit{essential graph}, which is less dense than the covisibility graph because an edge between two KFs is created when they share more than 100 landmark observations.
Finally, the map also contains a spanning tree, which is the minimum connected representation of a graph that includes all the KFs.

\textbf{Feature Tracking.} We perform feature tracking through the stereo visual odometry algorithm from our previous work \cite{gomez2016robust}.
In a nutshell, we track image features (points and segments) from a sequence of
stereo frames and compute their 3D position and their associated uncertainty represented by covariance matrices. 
The 3D landmarks are then projected to the new camera pose, and the projection errors are minimized in order to obtain both the camera pose increment and the covariance associated to such estimation.
This process is repeated every new frame, 
\ruben{performing simply frame to frame VO,}
until a new KF is inserted to the 
map. Further discussion about this feature tracking procedure will be formally addressed in \secref{sec_tracking}. 
Once a KF is inserted into the map, two procedures are run in parallel: local mapping and loop closure detection.

\textbf{Local Mapping.} The local mapping procedure looks for new feature correspondences between the new KF, the last one and those connected to the last one in the covisibility graph. 
This way, we build the so-called \textit{local map} of the current KF, which includes all the KFs that share at least 20 landmark observations with the current one as well as all the landmarks observed by them. 
Finally, an optimization of all the elements within the local map (KF poses and landmarks positions) is performed. 
A detailed description of this procedure will be presented in \secref{sec_mapping}.
%
	
\textbf{Loop Closure.} In parallel to local mapping, a loop closure detection is carried out by extracting a visual descriptor for each image, based on a bag-of-words approach, as will be described in \secref{sec_loop_closure}. 
All the visual descriptors of the captured frames during camera motion are stored in a database, which is later employed to find similar frames to the current one. 
The best match will be considered a loop closure candidate only if the local sequence surrounding this KF is also similar.
Finally, the relative $\symEucSpace$ transformation between the current KF and the loop closure candidate is estimated so that, if a proper estimation is found, all the KFs poses involved in the loop are corrected through a pose-graph optimization (PGO) process.

It is important to remark that the stereo visual odometry system runs continuously
at every frame while both the local mapping and loop closure detection procedures are launched in background 
(in separated threads) only when a new KF is inserted, thus allowing our system to reach real-time performance.
\review{
In the event of a new keyframe being inserted in the system	while the local mapping thread is still being processed, the keyframe is temporary stored until the map is updated and then a new local mapping process is launched.
}

\review{These mapping and loop closure approaches are identical to the ones followed in ORB-SLAM, 
being aimed to reduce the high computational burden that general BA involves (along with the incorporation of recent sparse algebra techniques). Within the BA framework, our proposal belongs to the so-called \textit{relative} techniques (e.g. \cite{sibley2009adaptive,sibley2010vast,moreno2016constant}), which have gained great popularity in the last years as an alternative to the more costly \textit{global} approaches (e.g. \cite{klein2007parallel,kaess2008isam}).
}

%
%
%

%

%% file: 4_feature_tracking.tex
\section{Feature Tracking}
\label{sec_tracking}
This section reviews the most important aspects of our previous work \cite{gomez2016robust}, which deals with the visual odometry estimation between 
consecutive frames, and also with the KF decision policy.
Basically, both points and line segments are tracked along a sequence of stereo frames (see \fig{fig_feature_tracking}), and then the 3D motion of the camera (and also its uncertainty) is computed by minimizing the projection errors.

\subsection{Point Features}
In this work we use the well-known ORB method \cite{rublee2011orb} due to its great performance for keypoint detection, and the binary nature of the descriptor it provides, which allows for a fast, efficient keypoint matching.
%
%
In order to reduce the number of outliers, we only consider measurements which fulfill that the best match in the left image corresponds to the best match in the right one, i.e. they are mutual best matches.
Finally, we also filter out those matches whose distance in the descriptor space with the second best match is less than twice the distance with the best match, to ensure that the correspondences are meaningful enough.

\subsection{Line Segment Features}
The Line Segment Detector (LSD) method \cite{von2010lsd} has been employed to extract line segments, providing high precision and repeatability.
For stereo matching and frame-to-frame tracking we augment line segments with a binary descriptor provided by the Line Band Descriptor (LBD) method \cite{zhang2013efficient}, which allows us to find correspondences between lines based on their local appearance.
%
%
Similarly to the case of points, we check that both candidate features are mutual best matches, and also that the feature is meaningful enough.
\review{
Finally, we take advantage of the useful geometrical information that line segments provide in order to filter out those line matches with different orientations and lengths, and those with a high difference on the disparities of the endpoint.
Notice that this filter helps the system to retain a larger amount of structural lines, which allows for the formation of more consistent
maps based on points and lines (see \fig{euroc-map}).
}

\subsection{Motion Estimation}
\label{sec_mot_est}
%
%
Once we have established the correspondences between two stereo frames, we back-project both the keypoints and the line segments from the first frame to the next one.
Then, we iteratively estimate the camera ego-motion through a robust Gauss-Newton minimization of the line and keypoint projection errors.
In order to deal with outliers, we employ a Pseudo-Huber loss function and perform a two-step minimization, as proposed in \cite{moreno2013erode}.
Finally, we obtain the incremental motion estimation between the two consecutive frames, which can be modelled by the following normal distribution:
\begin{equation}
\label{eq_motion}
\separam_{t,t+1} \sim \mathcal{N}(\separamopt_{t,t+1},\symcov_{\separamopt_{t,t+1}})
\end{equation}
where $\separamopt_{t,t+1} \in \symEucLie$ is the 6D vector of the camera motion between the frames $t$ and $t+1$, and $\symcov_{\separamopt_{t,t+1}}$ stands for the covariance of the estimated motion\review{, approximated by the inverse of
the Hessian of the cost function in the last iteration}.

\subsection{Keyframe Selection}
%
%
For deciding when a new KF is inserted in the map, we have followed the approach in \cite{kerl2013dense} which employs the uncertainty of the relative motion estimation.
Thus, following equation \eqref{eq_motion}, we transform the uncertainty from the covariance matrix into a scalar, named \textit{entropy}, through the following expression:
\begin{equation}
\label{eq_entropy}
h(\separam) = 3(1+log(2\pi)) + 0.5\:log(|\symcov_{\separam}|)
\end{equation}
%
Then, for a given KF $i$ we check the ratio between the entropy from the motion estimation between the previous KF $i$ and the current one $i+u$ and that between the previous KF $i$ and its first consecutive frame $i+1$, i.e.:
\begin{equation}
\alpha = \frac{h(\separam_{i,i+u})}{h(\separam_{i,i+1})}
\end{equation}
If the value of $\alpha$ lies below some pre-established threshold, which in our experiments has been set to 0.9, then the frame $i+u$ is inserted to the system as a new KF.
\review{
Notice that to compute the expression in Equation \eqref{eq_entropy}, we need the uncertainty of the pose increment between non-consecutive frames. Since Equation \eqref{eq_motion} only estimates the incremental motion between consecutive frames, a series of such estimations are composed through first order Gaussian propagation techniques to obtain the covariance between two non-consecutive KFs.}
%

%% file: 5_mapping.tex
\section{Local Mapping}
\label{sec_mapping}
This section describes the behavior of the system when a new KF is inserted, which essentially consists in performing the bundle adjustment of the so-called \textit{local map} i.e.: those KFs connected with the current one by the covisibility graph and the landmarks observed by those local KFs.

\subsection{Keyframe Insertion}
Every time the visual odometry thread selects a KF, we insert it into the SLAM system and optimize the local map. 
First, we refine the {estimation of the relative pose change} between the current and the previous KFs, since the one provided by the VO is estimated by composing the relative motions between the intermediate frames.
\review{
For that, we perform data association between the KFs, taking into account the geometrical restrictions described in \secref{sec_tracking} and obtaining a consistent set of common features observed in them. 
}
Then, we perform a similar optimization than the one presented in \secref{sec_mot_est}, for which we employ the pose provided by the VO thread as the initial estimation for a Gauss-Newton minimization.
Once we have computed the relative {pose change} between the KFs, we insert the current one into the system, including:
\begin{enumerate}
\item {An index for the keyframe.}
\item {The information of its 3D pose, which comprises an absolute pose and the relative pose from the previous KF, along with their associated uncertainties}.
\item The new 3D landmarks, which are initialized by storing both their 2D image coordinates and their descriptors. The {new} observations of the already existing landmarks are also {added to the map}.
\end{enumerate}
Finally, we also look for new correspondences between the unmatched feature observations from the current frame, and the landmarks in the local map.

\subsection{Local Bundle Adjustment}
\label{sec_lba}
After inserting the KF, the next step is to perform a bundle adjustment of the local map. As stated before, this map is formed by all the KFs connected with the current one in the covisibility graph (i.e. those that share at least 20 landmarks) and also all the landmarks observed by the local KFs.
For that, let us define the vector $\lbavar$ that contains the variables to be optimized, which are the $\symEucLie$ pose of each KF $\kf{i}$, the 3D position of each point $\pt{j}$, and also the 3D positions of the endpoints for each line segment: $\{ \lsPL{wk}, \lsQL{wk} \}$.
Then, we minimize the projection errors between the observations and the landmarks projected to the frames where they were observed:
\begin{equation}
\label{eq_lba}
\lbavar^* = \argmin{\lbavar} \sum\limits_{i \in \kfgroupL}
\bigg[
\sum\limits_{j \in \ptgroupL} \textbf{e}_{ij}\TRANSPOSE \symcov_{\textbf{e}_{ij}}^{-1}  \textbf{e}_{ij}
+
\sum\limits_{k \in \lsgroupL} \textbf{e}_{ik}\TRANSPOSE \symcov_{\textbf{e}_{ik}}^{-1}  \textbf{e}_{ik}
\bigg]
\end{equation}
where $\kfgroupL$, $\ptgroupL$ and $\lsgroupL$ refer to the groups of local KFs, points, and line segments, respectively.
%
%

In this expression, the projection error $\textbf{e}_{ij}$ stands for the 2D distance between the observation of the $j$-th map point in the $i$-th KF, and can be expressed as:
\begin{equation}
\textbf{e}_{ij} = \ptproj{ij} - \projfun{ ( \kf{i}, \pt{j} ) }
\end{equation}
where the function
$
\projfun{}: \symEucLie \cross \mathbb{R}^3 \mapsto \mathbb{R}^2
$
first places the \textit{j}-th 3D point $\pt{j}$ (in world coordinates) into the local reference system of the $i$-th KF, i.e. {$\ptL{ij}$}, and then projects this point to the image.
The use of line segments is slightly different, since we cannot simply compare the position of the endpoints as they might be displaced along the line or occluded from one frame to the next one. For that, we take {as error function} the distances between the projected endpoints of the 3D line segment and its corresponding infinite line in the image plane.
In this case, the error $\textbf{e}_{ik}$ between the $k$-th line observed in the $i$-th frame, is given by:
\begin{equation}
\textbf{e}_{ik} = 
\begin{bmatrix}
\lsproj{ik} \cdot \projfun{ ( \kf{i}, \lsPL{wk} ) } \\
\lsproj{ik} \cdot \projfun{ ( \kf{i}, \lsQL{wk} ) } \\
\end{bmatrix}
\end{equation}
where $\lsPL{wk}$ and $\lsQL{wk}$ refer to the 3D endpoints of the line segments in the world coordinate system and $\lsproj{ik}$ is the equation of the infinite line that corresponds to the $k$-th line segment in the $i$-th KF, which can be obtained with the cross product between the 2D endpoints of the line segments in homogeneous coordinates, i.e.: $\lsproj{ik} = \lsPproj{ik} \cross \lsQproj{ik}$. 

The problem in \eqref{eq_lba} can be iteratively solved by following the Levenberg-Marquardt optimization approach, for which we need to estimate both the Jacobian and the Hessian matrices:
\begin{equation}
\Delta \lbavar = \big[ \textbf{H} + \lambda \: diag(\textbf{H}) \big]^{-1} \textbf{J}\TRANSPOSE \textbf{W} \textbf{e}
\end{equation}
where the error vector $\textbf{e}$ contains all the projection errors $\textbf{e}_{ij}$ and $\textbf{e}_{ik}$.
This equation, along with the following update step:
\begin{equation}
\lbavar' = \lbavar \boxplus \Delta \lbavar
\end{equation}
\begin{figure*}[ht!]
\begin{center}
\begin{equation}
\label{eq_hessian}
\textbf{H} \approx 
%
\left[
\begin{array}{c|c|c}
\input{./submatrices/kk.tex} & \input{./submatrices/kp.tex} & \input{./submatrices/kl.tex} \\
\hline
\input{./submatrices/pk.tex} & \input{./submatrices/pp.tex} & \input{./submatrices/zz.tex} \\
\hline
\input{./submatrices/lk.tex} & \input{./submatrices/zz.tex} & \input{./submatrices/ll.tex}
\end{array}
\right]
\end{equation}
\end{center}
\end{figure*}
can be applied recursively until convergence, resulting in the optimal $\lbavar$, from which we can update the position of the local KFs and landmarks. Notice that the update equation cannot be directly applied to the whole vector, given the different nature of the variables in $\lbavar$.
It is important to remark that each observation error $\textbf{e}_{ij}$ or $\textbf{e}_{ik}$, only depends on a single KF $\kf{i}$, and a single landmark $\pt{j}$ or $\{ \lsPL{wk}, \lsQL{wk} \}$.
Hence, the Hessian matrix can be formed by appending the influence of each observation to its corresponding block, as showed in Equation \eqref{eq_hessian}.
%
%
Notice that, for the rest of observations that belong to the KFs that are not part of the local map, their Jacobian matrixes $\der{\textbf{e}_{ij}}{\kf{i}}$ and $\der{\textbf{e}_{ik}}{\kf{i}}$ are equal to zero, since here we only optimize the local map while the rest of the Kfs remain fixed. 

{It should also be underlined that in \eqref{eq_lba} the influence of the errors in both points and lines is weighted with $\symcov_{\textbf{e}_{ij}}^{-1}$ and $\symcov_{\textbf{e}_{ik}}^{-1}$, respectively, which stand for the inverses of the covariance matrixes associated to the uncertainty of each projection error.}
{In practice, though, it is more effective to set such covariances to the identity matrix and follow a similar approach to the one described in \secref{sec_mot_est} as it introduces robust weights and also deals with the presence of outlier observations.}

Finally, we remove from the map those landmarks with less than 3 observations, as they are less meaningful.

%% file: submatrices/kk.tex
\begin{matrix}
 & \hdots & \\
\vdots &
\textcolor{\ptcolor}{ 
\der{\textbf{e}_{ij}}{\kf{i}}\TRANSPOSE \symcov_{\textbf{e}_{ij}}^{-1} \der{\textbf{e}_{ij}}{\kf{i}}
}
+
\textcolor{\lscolor}{ 
\der{\textbf{e}_{ik}}{\kf{i}}\TRANSPOSE \symcov_{\textbf{e}_{ik}}^{-1} \der{\textbf{e}_{ik}}{\kf{i}}
}
 & \vdots \\
 & \hdots & 
\end{matrix}

%% file: submatrices/kp.tex
\begin{matrix}
 & \hdots &  \\
\vdots & 
\textcolor{\ptcolor}{ 
\der{\textbf{e}_{ij}}{\kf{i}}\TRANSPOSE \symcov_{\textbf{e}_{ij}}^{-1} \der{\textbf{e}_{ij}}{\pt{j}}
}
 & \vdots \\
 & \hdots & 
\end{matrix}

%% file: submatrices/kl.tex
\begin{matrix}
 & \hdots &  \\
\vdots & 
\textcolor{\lscolor}{ 
\der{\textbf{e}_{ik}}{\kf{i}}\TRANSPOSE \symcov_{\textbf{e}_{ik}}^{-1} \der{\textbf{e}_{ik}}{ \{ \lsP{k}, \lsQ{k} \} }
}
 & \vdots \\
 & \hdots & 
\end{matrix}

%% file: submatrices/pk.tex
\begin{matrix}
 & \hdots &  \\
\vdots & 
\textcolor{\ptcolor}{
\der{\textbf{e}_{ij}}{\pt{j}}\TRANSPOSE \symcov_{\textbf{e}_{ij}}^{-1} \der{\textbf{e}_{ij}}{\kf{i}} 
}
 & \vdots \\
 & \hdots & 
\end{matrix}

%% file: submatrices/pp.tex
\begin{matrix}
 & \hdots & \\
\vdots & 
\textcolor{\ptcolor}{
\der{\textbf{e}_{ij}}{\pt{j}}\TRANSPOSE \symcov_{\textbf{e}_{ij}}^{-1} \der{\textbf{e}_{ij}}{\pt{j}}
}
 & \vdots \\
 & \hdots & 
\end{matrix}

%% file: submatrices/zz.tex
\begin{matrix}
 & \hdots & \\
\vdots & 
\textbf{0}
 & \vdots \\
 & \hdots & 
\end{matrix}

%% file: submatrices/lk.tex
\begin{matrix}
 & \hdots &  \\
\vdots & 
\textcolor{\lscolor}{
\der{\textbf{e}_{ik}}{ \{ \lsP{k}, \lsQ{k} \} }\TRANSPOSE \symcov_{\textbf{e}_{ik}}^{-1} \der{\textbf{e}_{ik}}{\kf{i}}
}
 & \vdots \\
 & \hdots & 
\end{matrix}

%% file: submatrices/ll.tex
\begin{matrix}
 & \hdots &  \\
\vdots & 
\textcolor{\lscolor}{
\der{\textbf{e}_{ik}}{ \{ \lsP{k}, \lsQ{k} \} }\TRANSPOSE \symcov_{\textbf{e}_{ik}}^{-1} \der{\textbf{e}_{ik}}{ \{ \lsP{k}, \lsQ{k} \} }
}
 & \vdots \\
 & \hdots & 
\end{matrix}

%% file: 6_loop_closure.tex
\section{Loop Closure}
\label{sec_loop_closure}
In this work, we adopt a bag of words \paco{(BoW)} approach based on the binary descriptors extracted for both the keypoints and the line segments in order to \paco{robustly} cope with data association and loop closure detection. 

\paco{In short,} the BoW technique consists in summarizing all the information extracted from an image (in our proposal, the descriptors of keypoints and line segments) into a \textit{word} vector, \paco{employing for that a vocabulary that has been built off-line from different image datasets}. 
\paco{Then, as the camera moves, the words computed from the grabbed images are stored in a database that is later employed to seek for the most similar image to the current keyframe.}

\paco{In the following, we first address the process of detecting loop closures from the created BoWs, and then describe the correction of the pose estimations of the keyframes involved in the loop.}

\subsection{Loop Closure Detection}
\label{sec_lc_detection}
{
\paco{The detection of loop closures involves both to find an image similar to the one being currently processed and to estimate the relative pose change between them, as described next.}
}

\subsubsection{Visual Place Recognition}
Specifically, we have \paco{employed} the method presented in \cite{galvez2012bags}, which was initially developed for BRIEF binary descriptors, and subsequently adapted to ORB keypoints. 
Since, in our work, segments are also augmented with binary descriptors, we propose to build both specific visual vocabularies and databases for them. 
This way, at each time step, the most similar images in the databases of keypoints and segments are retrieved in parallel in order to look for loop closures. This dual-search is motivated by the fact that some scenes may be described more distinctively by segments than by keypoints or vice versa. Thus, employing both methods and merging their results allow us to refine the output of database queries, incurring in a small computational footprint.

To illustrate this, we first define a \textit{similarity matrix} as the matrix that contains in each row the similarity values, in the range [0,1], of a certain image with all the images stored in the database. \paco{Then, we compute such matrices from a sequence recorded in a corridor that goes around a square area}.

%
\begin{figure}[tb]
  \centering
    \includegraphics[width=\wconfmat]{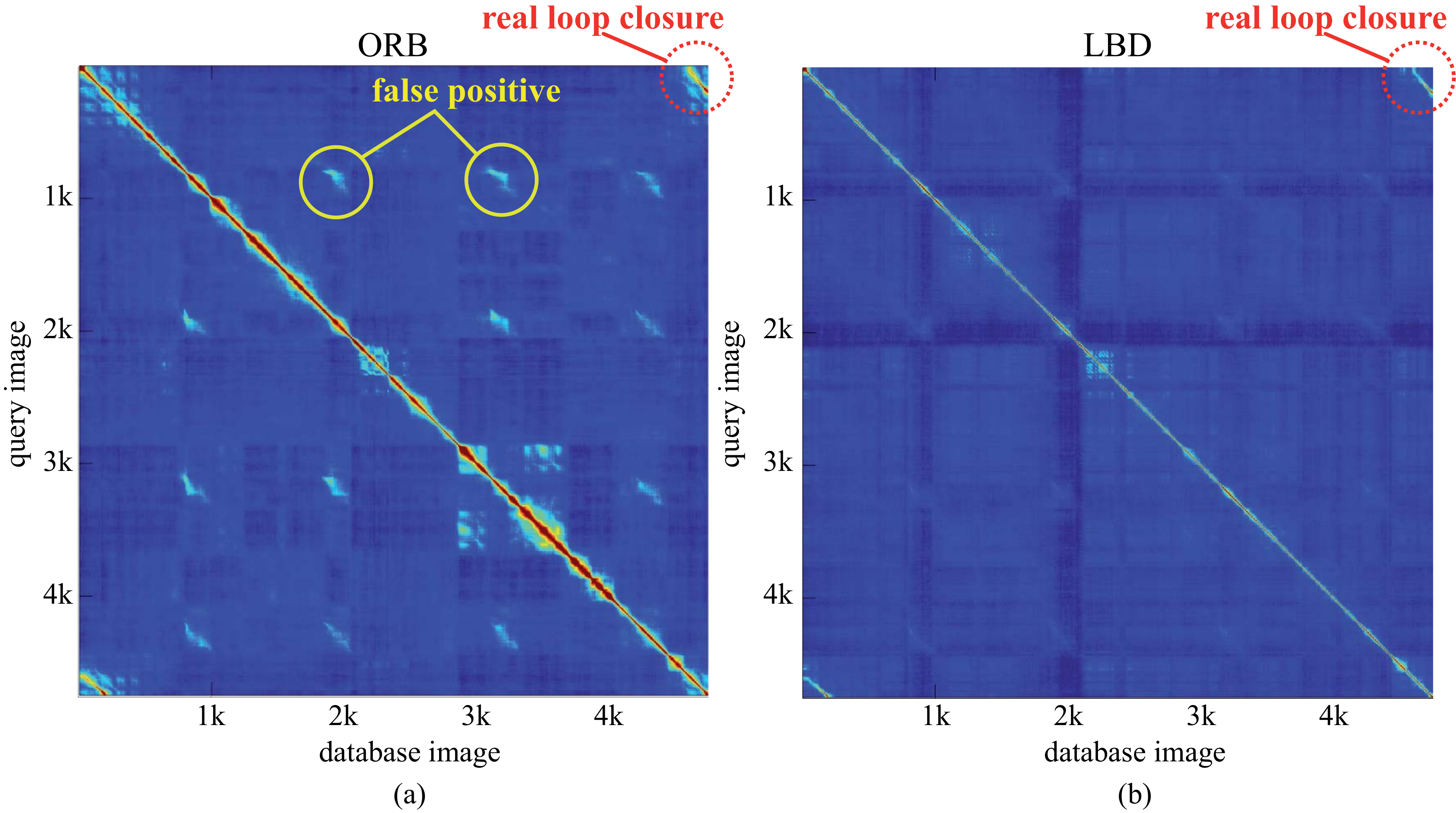}
    \caption{Similarity matrices for a certain dataset where the (a) ORB keypoint-only bag-of-words approach yields false positives that are not present in the (b) LBD line-only approach.}
    \vspace{\indfig}
    \label{fig_sim_mat}
\end{figure}
Concretely, the matrix in \fig{fig_sim_mat}(a) has been computed employing only ORB keypoints to build both the vocabulary and the database, while the other (\fig{fig_sim_mat}(b)) relies only on segments. The color palette goes from blue (score = 0) to red (score = 1). As can be noted, some yellowish areas appear in the first matrix in places where the images look similar according to the keypoints (specifically, after turning at the corners of the corridor). This indicates potential loop closures although, \paco{in fact}, they are \paco{just} false positives. The second (line-only) matrix, though, does not present this behavior so that it may be employed to discard them. On the other hand, the first matrix presents more distinctiveness, since the \paco{difference in} score is generally larger for non-similar images than in the line-only matrix. \paco{Therefore, the image similarities yielded by querying both feature databases may be combined to improve robustness when detecting potential loop closures}.

\paco
{
	In this work, we propose to weight the results from both features ($s_k$ for keypoints and $s_l$ for lines) according to two criteria, namely \textit{strength} and \textit{dispersion}. The former weights the similarity score proportionally to the number of features of a certain type (keypoint or line) in the set of features detected in the image, while the latter takes into account the dispersion of the features in the image (the more disperse the higher the weight will be). This yields a more robust total similarity score for the image ($s_t$):
}
\paco
{
	\begin{equation}
	\begin{split}
	s_t = 0.5\left(n_k/\left(n_k+n_l\right)+d_k/\left(d_k+d_l\right)\right)s_k + \\0.5\left(n_l/\left(n_k+n_l\right)+d_l/\left(d_k+d_l\right)\right)s_l,
	\end{split}
	\end{equation}
}
\paco
{
	where $n_k$ and $n_l$ are the number of keypoints and segments extracted in the image, respectively, and $d_k$ and $d_l$ are the dispersion values, which are computed as the square root of the sum of the variances in the $x$ and $y$ coordinates of the found features. For the case of segments, the midpoint coordinates are employed.
}
\review{Note that this formulation gives the same importance to both kinds of features (hence the 0.5 factor), although it could be tuned according to the environment (e.g. if the images are expected to be low-textured, it might be more convenient to down-weight the keypoint result with respect to the lines one).}
\paco
{
	We have empirically evaluated this strategy in comparison to four other alternatives following the classification framework employed in \cite{gomez2015training} for four different datasets: Oxford dataset \cite{smith2009new}, sequence 4 in Malaga dataset \cite{Blanco-Claraco2014}, sequence 7 in KITTI dataset \cite{geiger2012we} and i3tf dataset \cite{zhang2015building}. 
	The compared approaches were: i) using just $s_k$, ii) using just $s_l$, iii) using both $s_k$ and $s_l$ but taking into account only the \textit{strength} criteria, and iv) using both $s_k$ and $s_l$ but taking into account only the \textit{dispersion} criteria. Our proposed strategy yielded better results overall in terms of precision-recall for all the datasets.
}

\subsubsection{Estimating the Relative Motion}
Once we have a loop closure candidate, we still need to discard false positives that could have not been detected with the above mentioned approach. This is achieved by recovering the relative pose between the two KFs involved in the loop closure (namely \textit{current} and \textit{old} KFs from now on). For that, we first look for matches between the features from both KFs in a similar way to the one described in \secref{sec_overview}, while also searching for new correspondences between the current KF and the local map associated to the \textit{old} one.
Then, we estimate a valid transformation $\hat\separam_{ij} \in \symEucLie$ that relates both KFs following the approach described in \secref{sec_mot_est}. 
Finally, since an erroneous detection of a loop closure (false positive) would produce a very negative impact on the SLAM system, we check the consistency of the loop closure candidate with the following tests:
\begin{enumerate}[i)]
\item The maximum eigenvalue of the covariance matrix $\symcov_{\hat\separam_{ij}}$ is inferior to 0.01.
\item The obtained translation and rotation cannot rise over 0.50 meters and 3.00 degrees, respectively.
\item The inliers ratio in the estimation is higher than 50$\%$.
\end{enumerate}
\review{
Regarding the first criterion, a large value of the eigenvalues of the uncertainty matrix (see \eqref{eq_motion}) is often an indicator of an ill-conditioned Hessian matrix, most probably due to the presence of a large number of outliers in the feature matching set. Ensuring that the maximum eigenvalue of the covariance matrix is below a certain threshold allows us to detect potentially incorrect loop closures candidates and discard them.
%
}

\review{
	In the case of the second criterion, we also set a maximum translation and rotation limit for the estimated pose, as BoW-based approaches typically provides positive matches that are very similar in appearance and pose, so that a large change in pose between the involved frames usually indicates a wrong loop closure detection. Finally, the third criterion sets a minimum ratio of detected inliers after the optimization process, since motion estimation is strongly affected by the presence of outliers and incorrect associations from visual place recognition.
}


\subsection{Loop Correction}
After estimating all consecutive loop closures in our trajectory, we then fuse both sides of the loop closure correcting the error distributed along the loop.
This is typically solved by formulating the problem as a pose-graph optimization (PGO), where the nodes are the KFs inside the loop, and the edges are given by both the essential graph and the spanning tree.
For that, let us define the following error function as the $\symEucLie$ difference between the transformation that relates the KFs $\hat\separam_{ij}$ to the current observation of the same transformation:
\begin{equation}
\label{eq_loopclosure}
\textbf{r}_{ij} ( \kf{i}, \kf{j} ) = 
log
\big( 
exp(\hat\separam_{ij}) \cdot exp(\kf{j}) \cdot exp(\kf{i})^{-1}
\big)
\end{equation}
where the operators $log: \symEucSpace \mapsto \symEucLie$ and $exp: \symEucLie \mapsto \symEucSpace$  refer to the well-known logarithm and exponential maps.
Notice that in the case of a regular edge, the value of $\hat\separam_{ij}$ coincides with the estimation of $\separam_{ij}$ in the first step of the optimization, and hence the error in these edges is initially zero.

{
This PGO problem is solved using the g2o library \cite{kummerle2011g} yielding the optimal pose of the KFs included in the optimization, i.e. the essential graph and the spanning tree, when considering the loop closure edges.
}
%
%
Finally, we update the pose of the KFs along with the pose of the landmarks observed by them, and we also merge the local maps of both sides of the loop by first fusing the landmarks matched while estimating their relative motion (please, refer to  \secref{sec_lc_detection}), and then looking for new correspondences between the unmatched landmarks.

%% file: 7_experiments_2ndreview.tex
\newcommand\Tstrut{\rule{0pt}{2.6ex}}         
\newcommand\Bstrut{\rule[-0.9ex]{0pt}{0pt}}   
\section{Experimental Validation}
\label{sec_experiments}
In this section we evaluate the performance of PL-SLAM 
in several scenarios from different datasets, for which we estimate both the trajectory and the map in several video sequences. 
We also compare our approach with the stereo version of ORB-SLAM \cite{mur2016orb} by employing its open source implementation, which is {considered one of the} state-of-the-art methods {for} stereo visual SLAM.

All the experiments {have been} run on an Intel Core i5-6600 CPU @ 3.30GHz and 16GB RAM without GPU parallelization.
{In order to fairly compare} all the sequences, we {have only considered} the relative errors between the KFs positions, {disregarding the accuracy} of the absolute poses since it dramatically varies depending on {whether or not} the sequence presents loop closures.
\begin{figure}[tb]
  \centering
    \includegraphics[width=\wkittimap]{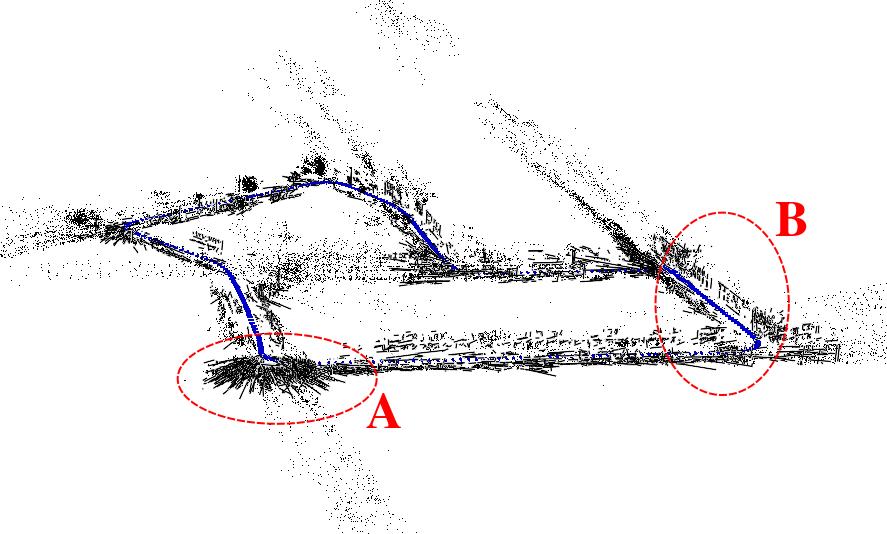}
    \caption{
    Map (in black) comprising points and line segments, and the trajectory (in blue) obtained with PL-SLAM from an outdoor {environment} in the sequence \textit{KITTI-07}. The map presents noisy 
    measurements in {some} parts {(e.g. zone A)}, and lines from the environment, such as parts of the buildings {(e.g. zone B)}. 
    }
    \vspace{\indfig}
    \label{fig_kitti_map}
\end{figure}
We {have} also tried to compare our method against the one proposed in \cite{zhang2015building}, but unfortunately, as {their approach to perform} line segment {tracking} is based on an optical flow algorithm, their proposal fails {when applied to} datasets with large motions between frames. Therefore, we could not include their results in this paper.

{In the following, we present examples of the trajectories and maps estimated by PL-SLAM, together with the average errors committed by our proposal, ORB-SLAM, a \textit{point-only} system (P-SLAM), and a \textit{line-only} system (L-SLAM).} 

\subsection{KITTI dataset}
First, we {have} tested PL-SLAM on the well-known KITTI dataset \cite{geiger2012we}, using the 11 sequences {that provide ground truth, yielding the results presented in \tab{tab_kitti}.}
Note that this is an urban dataset with highly textured image sequences and, as expected, the exploitation of line segments barely increases the accuracy, {since the number of detected points is generally sufficient for a proper operation of the SLAM system.}

{Still, PL-SLAM shows a slightly superior performance for most of the datasets in comparison to the \textit{point-only} approach and also to the ORB-SLAM system, specially in the rotation estimation.}
%
%
Unsurprisingly, the results confirm worse performance of the \textit{line-only} {system in these outdoor scenarios, even failing at properly estimating the trajectory of the stereo camera in some of the sequences (those recorded in rural environments).
}


As an illustrative example, \fig{fig_kitti_map} depicts the trajectory and the map estimated by PL-SLAM (LSD) in the sequence \textit{KITTI-07}.
As can be {seen} in 
{the zone marked as A in the figure,} the presence of line segments {can} introduce some 'noise' in the maps, as not all the detected lines 
{have a significant meaning,} i.e. some lines do not belong to structural parts of the {environment}.
{Nevertheless, in other parts of the sequence, relevant information of the scene structure has been correctly captured in the map. This can be observed 
{the zone marked as B in the figure,} where the buildings can be clearly noticed, leading to a descriptive representation of the scene.}
{On the contrary, the presence of noisy points in the map is less noticeable to the human eye, as they do not provide as much spatial information as line segments.} 
\begin{figure*}[tb]
	\centering
	\subfigure[]{
		\includegraphics[width=\wkittitraj]{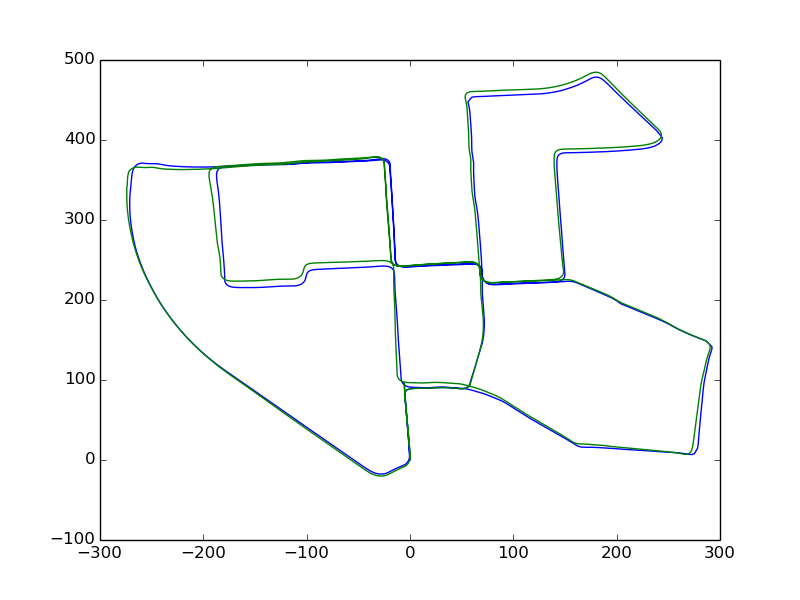}
		\vspace{\indfig}
		\label{kitti_x1}
		}
	~
	\subfigure[]{
		\includegraphics[width=\wkittitraj]{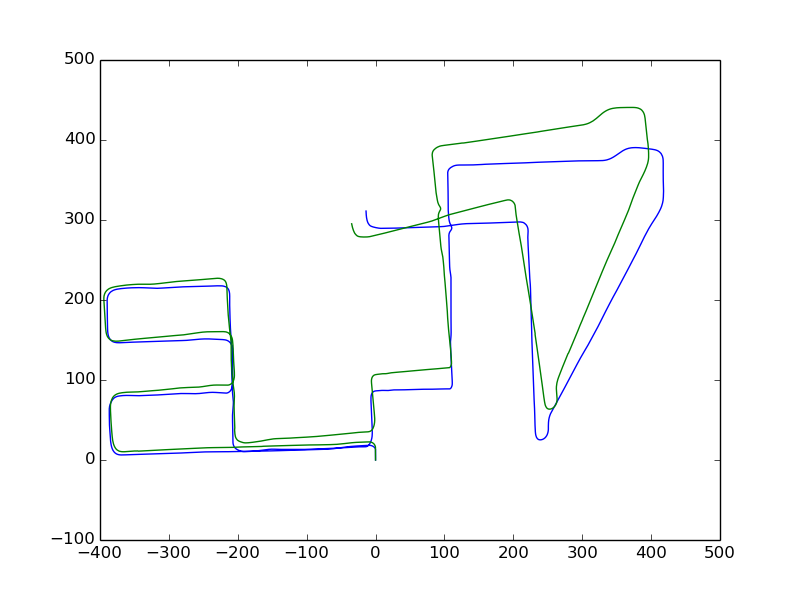}
		\vspace{\indfig}
		\label{kitti_x2}
		}
	\subfigure[]{
		\includegraphics[width=\wkittitraj]{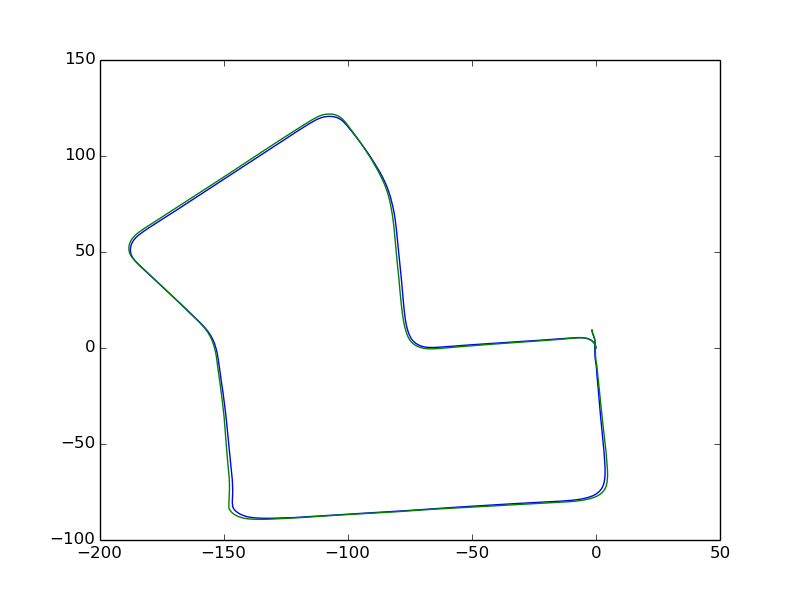}
		\vspace{\indfig}
		\label{kitti_x3}
		}
  	\caption{
  	Some trajectories estimated with PL-SLAM (in green) from the KITTI dataset (ground-truth in blue). 
  	\subref{kitti_x1} Trajectory estimated in the sequence \textit{KITTI-00}, where a large amount of loop-closures can be found.
  	\subref{kitti_x2} The sequence \textit{KITTI-08} does not present any loop closure, and hence the drift along the trajectory is not corrected.
  	\subref{kitti_x3} Finally, the sequence \textit{KITTI-07} presents a loop closure between the initial and final parts of the trajectory.
  	}
	\vspace{\indfig}
  	\label{fig_kitti_traj}
\end{figure*}

Finally, \fig{fig_kitti_traj} depicts the {estimated trajectory} obtained with PL-SLAM (LSD) in three sequences from the KITTI dataset that present different number of loop closures. It can be noted the importance of correcting the drift in long sequences to obtain accurate absolute solutions (refer to \fig{fig_kitti_traj}(a,c)), in contrast to the results obtained in sequences without loop closures, as the one presented in \fig{fig_kitti_traj}(b). Nevertheless, relative translation and rotation errors are similar for the three sequences, as shown in \tab{tab_kitti}.
\input{table_kitti_nofld.tex}
%

\subsection{EuRoC MAV dataset}
The EuRoC MAV dataset \cite{Burri25012016} consists of 11 stereo sequences recorded with a MAV flying across three different environments: two indoor rooms and one industrial scenario, containing sequences that present different challenges depending on the speed of the drone, illumination, texture, etc.
\begin{figure}[tb]
  \centering
	\subfigure[]{
		\includegraphics[width=\weurocmapa]{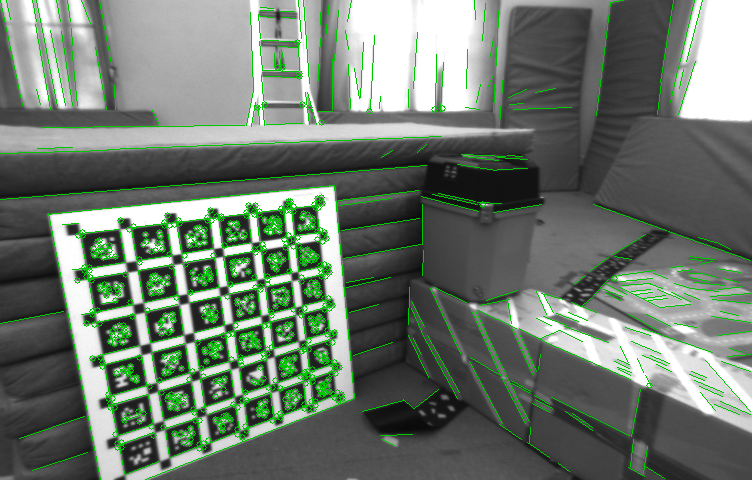}
		\vspace{\indfig}
		\label{euroc_img}
		}
	~
	\subfigure[]{
		\includegraphics[width=\weurocmapb]{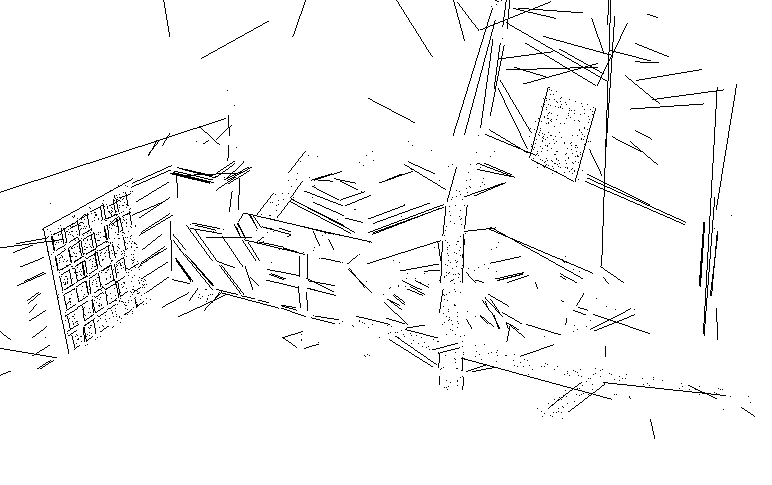}
		\vspace{\indfig}
		\label{euroc_map}
		}
  	\caption{
	Mapping results in the \textit{V1-01-easy} sequence from the EuRoC MAV dataset.
  	\subref{euroc_img} Features tracked between two consecutive keyframes.
  	\subref{euroc_map} {Resulting 3D map for the sequence. The checkerboard and the boxes in the scene are clearly reflected in the left part of the map, while more noisy features can be found in the rest, as a consequence of factors like non-textured surfaces, high illumination, etc.}  
  	}
    \vspace{\indfig}
    \label{fig_euroc_map}
\end{figure}
%
{As an example, we show the central part of the map built from the \textit{V1-02-easy} sequence in \fig{fig_euroc_map}(b), where two different parts are clearly visible.} The first one shows the features extracted from the non-structured part of the {environment (refer to the right side of the map)}, presenting a relatively large amount of small and noisy line segments, {which make} difficult the interpretation of that part of the {scene}.
{In contrast,} at the bottom left part of the figure, we can observe the structured part of the environment, which is clearly represented in the map through a set of line segments that {depicts} a checkerboard and a bunch of boxes. This example reflects that the maps built from line segments are {geometrically} richer than those created from only points, so that they can be employed to extract high-level meaningful information from them.

Finally, \tab{tab_euroc} shows the mean relative RMSE errors of the motion estimation in the different sequences included in the dataset. 
%
It can be observed that, for indoor and structured scenarios, {the inclusion of} line segment features in the system is very beneficial to estimate camera motion {and to improve the system robustness. In this case, both the \textit{point-only} and the \textit{line-only} approaches yield worse results than PL-SLAM, while ORB-SLAM fails in several sequences since feature tracking is prone to be lost.} 
%
\input{table_euroc_nofld.tex}
\if\hastum
\subsection{TUM dataset \cite{sturm12iros} }
Finally, we have tested our SLAM algorithm in the TUM dataset \cite{sturm12iros}, which is composed by a large number of sequences recorded with an RGB-D sensor. For that, we have adapted our approach to the case of RGB-D cameras by transforming the depth readings $Z_i$ to stereo disparity values $d_i$ through:
\begin{equation}
\label{eq_rgbd_stereo}
d_i = \frac{b \cdot f }{Z_i}
\end{equation}
where the baseline has been set to one for simplicity.
It is worth noticing that, apart from this, the rest of the system does not vary so that it still can be considered as a stereo visual system.

As an example, \fig{fig_tum} shows the clear and meaningful map obtained with our PL-SLAM approach for the sequence \textit{fr2-xyz}, which represents a very structured scenario.
The performance of our method has been further evaluated in several sequences from the TUM dataset, as shown in \tab{tab_tum}. 
It can be noted that the obtained results with our methods behave similarly to those for the previous dataset, with the L-SLAM version even outperforming the combined approach in some sequences as line segments are less suitable to present outliers in structured scenarios.
Again, the keypoint-only version is unable to finish the estimation of the camera trajectory due to the loss of keypoint tracking and the presence of outliers.
On the other hand, we observe the superior performance of ORB-SLAM in 4 of the tested sequences due to the benefits of such BA-oriented front-end in textured sequences with slow motions.
\begin{figure}[tb]
	\centering
	\subfigure[]{
		\includegraphics[width=\wtummap]{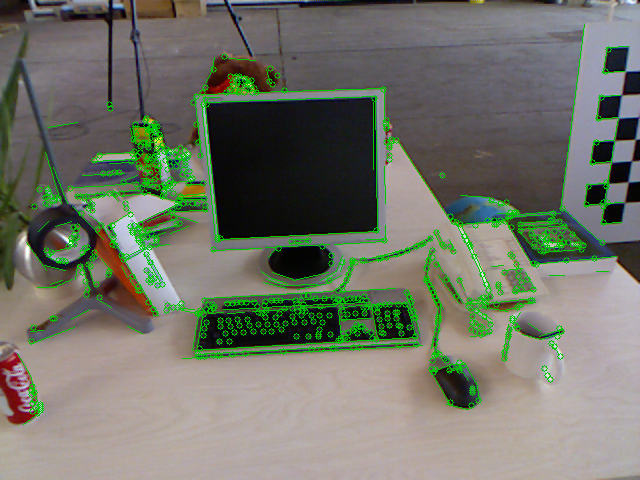}
		\vspace{\indfig}
		\label{tum_img}
		}
	~
	\subfigure[]{
		\includegraphics[width=\wtummap]{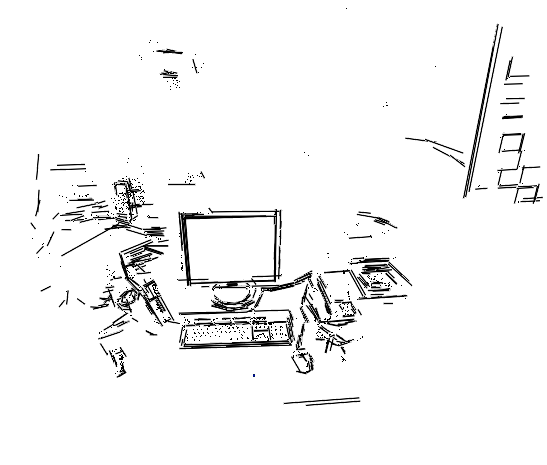}
		\vspace{\indfig}
		\label{tum_map}
		}
  	\caption{
  	Mapping results in the \textit{fr2-xyz} sequence from the TUM dataset. 
  	\subref{tum_img} Features tracked on a single keyframe.
  	\subref{tum_map} 3D map obtained with the first part of the sequence, where we can observe several informative line segments as a result of the line mapping in a structured environment.
  	}
	\vspace{\indfig}
  	\label{fig_tum}
\end{figure}
\begin{table}[tb]
\centering
\caption[]{Relative RMSE errors in the TUM dataset \cite{sturm12iros}. {A dash indicates that the experiment failed.}}
\label{tab_tum}
\resizebox{\columnwidth}{!}{%
\begin{tabular}{l|cccc}
Sequence & {P-SLAM} &  {L-SLAM} & {PL-SLAM} & {ORB-SLAM2} \\ 
\hline
\hline
fr1-desk	  	  & 0.xxxx  & 0.xxxx    & 0.xxxx    & 0.0420 \\
fr1-desk2   	  & 0.xxxx  & 0.xxxx    & 0.xxxx    & -      \\
fr1-room	 	  & 0.xxxx  & 0.xxxx    & 0.xxxx    & 0.0711 \\
fr2-xyz			  & 0.xxxx  & 0.xxxx    & 0.xxxx    & 0.0244 \\
fr3-longof 	      & 0.xxxx  & 0.xxxx    & 0.xxxx    & 0.0248 \\
fr3-ns-tn-wl	  & 0.xxxx  & 0.xxxx    & 0.xxxx    & 0.0405  
\vspace{\indtab}
\end{tabular}
}
\end{table}
\fi

\subsection{Low-textured Scenarios}
Finally, we {have assessed} the performance of the compared methods in challenging low-textured scenarios. For that, we have recorded a set of stereo sequences in a room {equipped with} an OptiTrack {system\footnote{http://optitrack.com/}, which provides the ground-truth of the camera trajectory.}
The resulting covisibility graph yielded by our PL-SLAM (LSD) system for the sequence \textit{lt-medium} is shown in \fig{fig_euroc_graph}, where a loop closure between the initial and the final part of the trajectory {can be observed}.
{These experiments (see \tab{tab_lowtext}) reveal that, while point-based approaches either fail to recover the trajectory or yield worse results than in previous scenarios, the two methods based on line segments are capable of robustly estimating the camera path, achieving a good performance in terms of accuracy.}
%
\input{table_lowtexture_nofld.tex}
%

\subsection{Performance}
Finally, regarding the computational burden, we present \tab{tab_runtime} that shows the average processing time of each part of the PL-SLAM algorithm, for each of the tested datasets.
\review{
Thanks to the efficient implementation of \cite{gomez2016robust} our VO thread achieves real-time performance 
for all combinations of features (i.e. points, lines, and points and lines) for the datasets with lower resolution 
(752$\times$480), and nearly real time in the KITTI dataset, and in all cases our approach performs faster 
than ORB-SLAM2, even when considering the two different features.
On the other hand, the local bundle adjustment (LBA) can be processed at around 20 Hz, which is fast enough for our purposes, as it runs in a parallel thread while the VO thread is continuously processing new frames.
Finally, it can be seen that the {loop closure management} is the most time consuming step of the algorithm {although} it is computed in a parallel thread (and not at every frame), so that the rest of the system can still run in real time.
} 
%
\input{table_perf_nofld.tex}
\subsection{Discussion}
\review{
As our system architecture is similar to in ORB-SLAM \cite{murTRO2015} we would like to clarify the essential differences between the two approaches: i) the inclusion of line segments as image features, which allows us to achieve robust camera localization in scenarios where keypoint-only methods usually perform poorly or even fails, ii) the inclusion of binary line descriptors in the loop closure procedure, in order to make it more robust, and iii) the implementation of the visual odometry thread as a frame-to-frame incremental motion estimation to meet the computational constraints that line segments introduces, unlike ORB-SLAM2, which performs motion-only Bundle Adjustment between recent frames.
In any case, we do not claim to obtain more accuracy than ORB-SLAM2 in common environments. 
In fact, both ORB-SLAM2 and our approach perform similarly in such environments, with slightly superior results for ORB-SLAM2 in accuracy and for our approach in computational burden. 
In essence, we claim more robustness in low-textured ones, where the number of point features dramatically decreases. Is in this kind of scenarios where our proposal achieves better performance.
}

\review{
Finally, although it was mentioned in the previous section, it is important to highlight that the results achieved by ORB-SLAM2 in these experiments have been computed by evaluating keyframe-to-keyframe pose estimation errors, which is a different metric than the one employed in the ORB-SLAM2 original paper, hence the differences in the results.
}


%% file: table_kitti_nofld.tex
\begin{table}[tb]
\centering
\caption[]{{Mean} results in the KITTI dataset \cite{geiger2012we}. The translation errors are expressed in $\%$, while the rotation errors are also expressed relatively to the translation in $deg/100m$. {A dash indicates that the experiment failed.}}
\label{tab_kitti}
\resizebox{\columnwidth}{!}{%
\begin{tabular}{l|cc|cc|cc|cc}
& \multicolumn{2}{c|}{P-SLAM} &  \multicolumn{2}{c|}{L-SLAM} & \multicolumn{2}{c|}{PL-SLAM} & \multicolumn{2}{c}{ORB-SLAM2}\\ 
Seq. & $t_{rel}$ & $R_{rel}$ & $t_{rel}$ & $R_{rel}$ & $t_{rel}$ & $R_{rel}$ & $t_{rel}$ & $R_{rel}$\\
\hline
\hline
00    & 2.57 & 3.00   & 3.29 & 7.99       & \textbf{2.38} & \textbf{2.32}   & 2.51 & 5.30 \\
01    & 5.59 & 2.32   & -    & - 	      & 3.23 & 2.17   & \textbf{1.51} & \textbf{1.65} \\
02    & 2.34 & 1.97   & 6.23 & 12.35      & \textbf{2.20} & \textbf{1.86}   & 2.31 & 3.88 \\
03    & 3.68 & 2.96   & 6.33 & 19.17      & 3.40 & \textbf{3.17}   & \textbf{3.31} & 5.52 \\
04    & 2.30 & 1.16   & -    & -          & 1.57 & \textbf{1.10}   & \textbf{1.45} & 2.21 \\
05    & 1.94 & 2.25   & 2.58 & 7.07       & \textbf{1.67} & \textbf{1.85}   & 1.75 & 4.52 \\
06    & 2.48 & 1.76   & 3.81 & 9.71       & 2.02 & \textbf{1.28}   & \textbf{1.45} & 2.86 \\
07    & 2.46 & 3.82   & 2.71 & 6.71       & \textbf{1.57} & \textbf{2.60}   & 2.20 & 5.43 \\
08    & \textbf{2.31} & \textbf{2.61}   & 6.97 & 13.67      & 2.42 & 2.65   & 2.50 & 5.03 \\
09    & 1.57 & \textbf{1.99}   & 6.56 & 11.55      & \textbf{1.49} & 2.12   & 1.63 & 3.95 \\
10    & 1.64 & 2.80   & 5.82 & 11.77      & \textbf{1.61} & \textbf{2.79}   & 1.81 & 6.43 
\end{tabular}
\vspace{\indtab}
}
\end{table}
%

%% file: table_euroc_nofld.tex
%
\begin{table}[tb]
\centering
\caption[]{Relative RMSE errors in the EuRoC MAV dataset \cite{Burri25012016}. {A dash indicates that the experiment failed.}}
\label{tab_euroc}
\resizebox{\columnwidth}{!}{%
\begin{tabular}{l|cccc}
Sequence & {P-SLAM} &  {L-SLAM} & {PL-SLAM} & {ORB-SLAM2} \\ 
\hline
\hline
MH-01-easy  	  & 0.0811  & 0.0588    & 0.0416             & \textbf{0.0251} \\
MH-02-easy  	  & 0.1041  & 0.0566    & \textbf{0.0522}    & 0.0638 \\
MH-03-med	      & 0.0588  & \textbf{0.0371}    & 0.0399    & 0.0712 \\
MH-04-dif         & -       & 0.1090    & 0.0641    & \textbf{0.0533} \\
MH-05-dif         & 0.1208  & 0.0811    & 0.0697    & \textbf{0.0414} \\
V1-01-easy  	  & 0.0583  & 0.0464    & 0.0423    & \textbf{0.0405} \\
V1-02-med  	      & 0.0608  & -         & \textbf{0.0459}    & 0.0617 \\
V1-03-dif	      & 0.1008  & -         & \textbf{0.0689}    & - \\
V2-01-easy  	  & 0.0784  & 0.0974    & \textbf{0.0609}    & - \\
V2-02-med  	      & 0.0767  & -         & \textbf{0.0565}    & 0.0666 \\
V2-03-dif	      & 0.1511  & -         & \textbf{0.1261}    & -   
\end{tabular}
\vspace{\indtab}
}
\end{table}

%% file: table_lowtexture_nofld.tex
%
\begin{table}[tb]
\centering
\caption[]{Relative RMSE errors in low-textured sequences recorded with GT data from an OptiTrack system. {A dash indicates that the experiment failed.}}
\label{tab_lowtext}
\resizebox{\columnwidth}{!}{%
\begin{tabular}{l|cccc}
Sequence & {P-SLAM} &  {L-SLAM} & {PL-SLAM} & {ORB-SLAM2} \\ 
\hline
\hline
lt-easy 	      & -  & 0.1412    & \textbf{0.1243}    & 0.1391 \\
lt-medium   	  & -  & 0.1998    & \textbf{0.1641}    & - \\ 
lt-difficult   	  & -  & 0.1801    & \textbf{0.1798}    & - \\
lt-rot-difficult  & 0.2411  & 0.2247   & \textbf{0.2034}    & 0.2910
\end{tabular}
\vspace{\indtab}
}
\end{table}

%% file: table_perf_nofld.tex
%
\begin{table}[tb]
\centering
\caption[]{Average runtime of each part of the algorithm.}
\label{tab_runtime}
\resizebox{\columnwidth}{!}{
\begin{tabular}{l|rrr}
 & \multicolumn{1}{c}{KITTI} & \multicolumn{1}{c}{EuRoC MAV} & \multicolumn{1}{c}{Low-Textured} \\
VO & \multicolumn{1}{c}{$1241\times376$} & \multicolumn{1}{c}{$752\times480$} & \multicolumn{1}{c}{$752\times480$} \\
\hline
\hline \Tstrut
P-SLAM								& 12.2 ms &  8.7 ms  &  8.1 ms \\
L-SLAM 							  	& 54.6 ms & 47.6 ms  & 46.1 ms \\
PL-SLAM 		    				& 66.0 ms & 49.7 ms  & 40.0 ms \\
ORB-SLAM2							& 98.1 ms & 69.0 ms  & 61.4 ms \\
\Tstrut
Local Mapping &  &  &  \\
\hline
\hline \Tstrut
P-SLAM								& 38.9 ms   &  37.3 ms  & 35.8 ms \\
L-SLAM 							  	& 37.4 ms   &  36.0 ms  & 34.5 ms \\
PL-SLAM 		    				& 43.8 ms   &  40.6 ms  & 42.1 ms   \\
ORB-SLAM2							& 230.0 ms  & 162.0 ms  & 102.0 ms   \\
\Tstrut
Loop Closing &  &  &  \\
\hline
\hline \Tstrut
P-SLAM								& 11.3 ms  & 3.5 ms  & 3.7 ms \\
L-SLAM 							  	& 9.5 ms   & 3.9 ms  & 3.4 ms \\
PL-SLAM 		    				& 28.0 ms  & 4.7 ms  & 4.5 ms   \\
ORB-SLAM2							& 9.1 ms   & 3.6 ms  & 4.4 ms 
\end{tabular}
\vspace{\indtab}
}
\end{table}

%% file: 8_conclusions.tex
\section{Conclusions}
\label{sec_conclusions}
In this paper we have proposed a novel stereo visual SLAM system that extends our previous VO approach in \cite{gomez2016robust}, and that is based on the combination of both keypoints and line segment features. 
Our proposal, {coined PL-SLAM,} contributes with a robust and versatile system capable of working in all types of environments, {including low-textured ones, while producing geometrically meaningful maps.}
{For that, we have developed the first open source SLAM system that runs in real time and that simultaneously employs keypoints and line segment features. Our ad-hoc implementation has been developed from scratch and its based on a bundle adjustment solution that seamlessly deals with the combination of different kinds of features.}
Moreover, we have extended the place recognition bag-of-words approach in \cite{galvez2012bags} for the case of simultaneously employing points and line segments {in order to enhance the loop-closure process.}
Our approach has been tested on popular benchmarking datasets such as KITTI, or {EuRoC MAV, as well as in a sequence of stereo images recorded in a challenging low-textured scenario. PL-SLAM has been compared with ORB-SLAM \cite{murTRO2015}, a point-only system and a line-only system, obtaining superior performance in terms of both accuracy and robustness in most of the dataset sequences.}

For future work, our implementation can benefit from better keypoint front-ends, such as the ones in SVO \cite{forster2014svo,forster2016svo} and PL-SVO \cite{gomez2016pl}, where authors reduced the computational time of the feature tracking with a semi-direct approach that estimates the position of the features as a consequence of the motion estimation.
%
Finally, our algorithm can be employed to obtain more accurate and refined maps by applying some SfM or Multi-Stereo techniques \cite{pizzoli2014remode,hofer2015line3d} in order to filter the structural lines, {hence} obtaining {more meaningful} information of the structured parts of the environment.

%

%% file: 9_bib.tex
\nocite{}
\bibliographystyle{ieeetr}
\bibliography{./biblio/plslam_biblio.bib}
%
\begin{IEEEbiography}[{\includegraphics[width=1in,height=1.25in,clip,keepaspectratio]{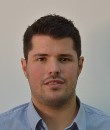}}]{Ruben Gomez-Ojeda}
(1988, Spanish) is a PhD student associated with the Machine Perception and 
Intelligent Robotics (MAPIR) group at the University of Malaga, under the
supervision of J. Gonzalez-Jimenez. He received his B.S.-M.S. in Industrial 
Engineering in 2012, and his M.S. in Mechatronics in 2014 from the University 
of Malaga, Spain. In 2016, he was a Visiting Researcher at the 
Robotics and Perception Group from the University of Zurich with D.Scaramuzza.
His research interests include vision based navigation, place recognition, and 
autonomous robotics.
\end{IEEEbiography}
\begin{IEEEbiography}[{\includegraphics[width=1in,height=1.25in,clip,keepaspectratio]{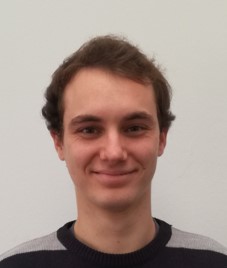}}]
	{David Zu\~niga-No\"el}
(1993, Spanish) received his B.S. degree in Computer Science in 2016 and his M.S. in Mechatronics in 2017,
both of them obtained at the University of Malaga, Spain. In 2016 he joined the Machine Perception and Intelligent
Robotics (MAPIR) group, where he started to work toward the PhD degree in Computer Vision and Robotics.
His research interests include vision based navigation, autonomous robotics, and sensor fusion. 
\end{IEEEbiography}
\begin{IEEEbiography}[{\includegraphics[width=1in,height=1.25in,clip,keepaspectratio]{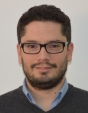}}]{Francisco-Angel Moreno}
(1981, Spanish) received his B.S. degree in Technical Telecommunications Engineering from the
University of Jaen in 2002. He received his M.S. degree in Telecommunications Engineering from 
the University of Malaga in 2007. In 2009 he joined the MAPIR group where he received his
PhD degree in 2015 under the supervision of Javier Gonzalez-Jimenez and Jose-Luis Blanco.
During his PhD he did two research stays, the first one in 2010 at the University of Bristol, and the 
second one in the University of Lincoln in 2013.
His research interests include vision based navigation, telepresence robotics and human-machine
interaction.
\end{IEEEbiography}
\begin{IEEEbiography}[{\includegraphics[width=1in,height=1.25in,clip,keepaspectratio]{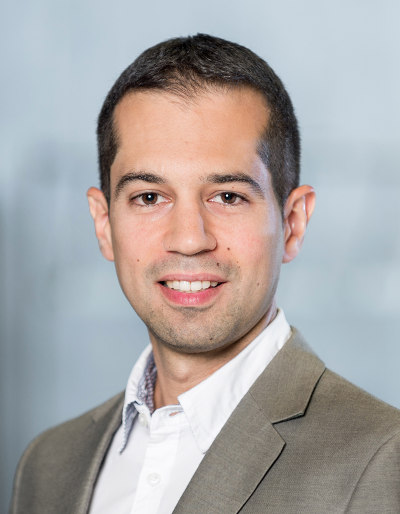}}]{Davide Scaramuzza}
(1980, Italy) 
received the Ph.D. degree in robotics and computer
vision from ETH Zurich, Zurich, Switzerland, in
2008, and a Postdoc at University of Pennsylvania,
Philadelphia, PA, USA.
He is a Professor of Robotics with University
of Zurich, where he does research at the intersection of 
robotics, computer vision, and neuroscience. From
2009 to 2012, he led the European project sFly, which
introduced the world's first autonomous navigation of
microdrones in GPS-denied environments using
visual-inertial sensors as the only sensor modality. He coauthored the book
Introduction to Autonomous Mobile Robots (MIT Press).
Dr. Scaramuzza received an SNSF-ERC Starting Grant, the IEEE Robotics
and Automation Early Career Award, and a Google Research Award for his
research contributions.
\end{IEEEbiography}
\begin{IEEEbiography}[{\includegraphics[width=1in,height=1.25in,clip,keepaspectratio]{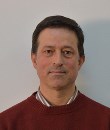}}]{Javier Gonzalez-Jimenez}
(1962, Spanish) is the head of the MAPIR group and full professor at the University of Malaga.
Prof. Gonzalez-Jimenez received his B.S. degree in Electrical Engineering from the 
University of Seville in 1987. He joined the Department of "Ingenieria de Sistemas y 
Automatica" at the University of Malaga in 1988 and received his Ph.D. from this 
University in 1993. In 1990-1991 he was at the Field Robotics Center, Robotics 
Institute, Carnegie Mellon University (USA) working on mobile robots as part
of his PhD. Since 1996 he has been heading Spanish and European projects on mobile 
robotics and perception. In these areas he is (co)author of more than 50 JCR-ISI 
papers, 100 international conferences and 3 books.
\end{IEEEbiography}